\documentclass{article}

\PassOptionsToPackage{numbers, compress}{natbib}

\usepackage[preprint]{neurips_2025}

\usepackage[utf8]{inputenc} %
\usepackage[T1]{fontenc}    %
\usepackage{hyperref}       %
\usepackage{url}            %
\usepackage{booktabs}       %
\usepackage{amsfonts}       %
\usepackage{nicefrac}       %
\usepackage{microtype}      %
\usepackage[table]{xcolor}         %

\usepackage{amsmath}
\usepackage{amsthm}
\usepackage{amssymb}
\usepackage{booktabs}
\usepackage{bm}
\usepackage{csquotes}
\usepackage{enumitem}
\usepackage{nicefrac}
\usepackage{pgfplots}
\usepackage{subcaption}
\usepackage{todonotes}
\usepackage{comment}
\usepackage{wrapfig}
\usepackage{siunitx}
\usepackage{todonotes}
\usepackage{hyperref}
\usepackage{url}

\usepackage{tikz}
\usepackage{graphicx}
\usetikzlibrary{arrows.meta, positioning, calc, patterns}

\newtheorem{proposition}{Proposition}

\renewcommand{\cite}[1]{\citep[][]{#1}}
\newcommand{\egcite}[1]{\citep[e.g.,][]{#1}}

\newcommand{\pspace}{\textbf{\textup{PSPACE}}}

\definecolor{mycolor0}{rgb}{0.12156862745098039, 0.4666666666666667, 0.7058823529411765}
\definecolor{mycolor1}{rgb}{1.0, 0.4980392156862745, 0.054901960784313725}
\definecolor{mycolor2}{rgb}{0.17254901960784313, 0.6274509803921569, 0.17254901960784313}
\definecolor{mycolor3}{rgb}{0.8392156862745098, 0.15294117647058825, 0.1568627450980392}
\definecolor{mycolor4}{rgb}{0.5803921568627451, 0.403921568627451, 0.7411764705882353}
\definecolor{mycolor5}{rgb}{0.5490196078431373, 0.33725490196078434, 0.29411764705882354}
\definecolor{mycolor6}{rgb}{0.8901960784313725, 0.4666666666666667, 0.7607843137254902}
\definecolor{mycolor7}{rgb}{0.4980392156862745, 0.4980392156862745, 0.4980392156862745}
\definecolor{mycolor8}{rgb}{0.7372549019607844, 0.7411764705882353, 0.13333333333333333}
\definecolor{mycolor9}{rgb}{0.09019607843137255, 0.7450980392156863, 0.8117647058823529}
\definecolor{darkNavy}{HTML}{003049}
\definecolor{specialRed}{HTML}{d62828}
\definecolor{specialOrange}{HTML}{f77f00}
\definecolor{specialYellow}{HTML}{fcbf49}

\title{Iterative Deployment Improves\\Planning Skills in LLMs}

\author{Augusto B. Corr\^{e}a\\
University of Oxford\\
United Kingdom
 \And
Yoav Gelberg\\
University of Oxford\\
United Kingdom
\And
Luckeciano C. Melo\\
University of Oxford\\
United Kingdom
\AND
Ilia Shumailov\\
AI Sequrity Company\\
United Kingdom
\And
Andr\'{e} G. Pereira\\
UFRGS\\
Brazil
 \And
Yarin Gal\\
University of Oxford\\
United Kingdom}

\begin{document}

\maketitle

\begin{abstract}

We show that iterative deployment of large language models (LLMs), each fine-tuned on data carefully curated by users from the previous models' deployment, can significantly change the properties of the resultant models.
By testing this mechanism on various planning domains, we observe substantial improvements in planning skills, with later models displaying emergent generalization by discovering much longer plans than the initial models.
We then provide theoretical analysis showing that iterative deployment effectively implements reinforcement learning (RL) training in the outer-loop (i.e.\ not as part of intentional model training), with an implicit reward function.
The connection to RL has two important implications: first, for the field of AI safety, as the reward function entailed by repeated deployment is not defined explicitly, and could have unexpected implications to the properties of future model deployments. Second, the mechanism highlighted here can
be viewed as an alternative training regime to explicit RL, relying on data curation rather than explicit rewards.
\end{abstract}

\section{Introduction}

In this paper, we show that repeatedly deploying large language models (LLMs)
and fine-tuning them on curated data from earlier deployments significantly
improves their planning capabilities.  This curation can be simply done by
\emph{validating} traces from previous generations, and \emph{selecting}
appropriate valid traces for future training. This mechanism produces a training
process conceptually similar to RL fine-tuning but where the reward signal is
left implicit.
The core idea is simple:
repeated deployment starts by users generating text with an LLM after its release. These texts go through a curation process, e.g.\ texts that do not capture user intent are rejected. Remaining texts are then shared to the web, with scrapes of the web including the curated text used to fine-tune the next generation of the LLM.

Iterative deployment is not a contrived setting: GPT-3.5 was trained on data
scraped from the web following GPT-3's deployment, with data shared on the web
at the time including curated texts generated by users using GPT-3
\cite{brown-et-al-neurips2020}. Similarly, GPT-4 was trained on data shared by
users from GPT-3.5 and GPT-3 \cite{openai-arxiv2024}, and so on. With agent
workflows becoming more commonplace, future training data will include agent
traces from prior model generations, similarly leading to an iterative training
on previous generation data. Figure~\ref{fig:pipeline} illustrates the
basics of this mechanism.

\input{method-figure}

\pgfplotsset{compat=1.11,
    /pgfplots/ybar legend/.style={
    /pgfplots/legend image code/.code={%
       \draw[##1,/tikz/.cd,yshift=-0.25em]
        (0cm,0cm) rectangle (10pt,0.8em);},
   },
}

\begin{figure}[t]
\centering
\begin{tikzpicture}
  \begin{axis}[
   ybar=0pt,
   bar width=20pt,
   width=13cm,
   height=4cm,
   enlarge x limits=0.35,
   enlarge y limits={upper,value=0.08},
   ymin=0,
   ymax=220,
   ylabel={\# solved tasks},
   symbolic x coords={Blocksworld,Rovers,Sokoban},
   xtick=data,
   xtick align=inside,
   x tick label style={font=\small},
   xtick style={draw=none},
   ytick style={draw=none},
   axis x line*=bottom,
   axis y line*=left,
   major y tick style = {black},
   grid=major,
   grid style={dashed,gray!30},
   legend style={at={(0.5,-0.18)}, anchor=north,legend columns=4, /tikz/every even column/.append style={column sep=6pt}},
   ylabel style={yshift=-2mm},
   cycle list name=exotic,
   nodes near coords,
   nodes near coords style={font=\footnotesize, /pgf/number format/fixed},
   every node near coord/.append style={yshift=2pt},
    ]

  \addplot[
    pattern=north east lines,
    pattern color=darkNavy,
    draw=black,
  ] coordinates {
    (Blocksworld,52)
    (Rovers,41)
    (Sokoban,32)
  };
  \addplot[draw=black, fill=specialRed] coordinates {
    (Blocksworld,109)
    (Rovers,131)
    (Sokoban,58)
  };
  \addplot[draw=black, fill=specialOrange] coordinates {
    (Blocksworld,132)
    (Rovers,201)
    (Sokoban,72)
  };

  \addplot[draw=black, fill=specialYellow] coordinates {
    (Blocksworld,154)
    (Rovers,205)
    (Sokoban,96)
  };

\legend{
  Base,
  Gen. 1,
  Gen. 2,
  Gen. 5
}

  \end{axis}
\end{tikzpicture}
\caption{Summary of our main results. Number of solved tasks (with 1000 tasks
  per domain) for three different domains when comparing the base model with
  later deployed generations (generations 1, 2, and 5). Average over three
  separate runs. In all domains, the fifth generation more than doubles the
  performance of the base model.}
\label{fig:results-summary}
\end{figure}
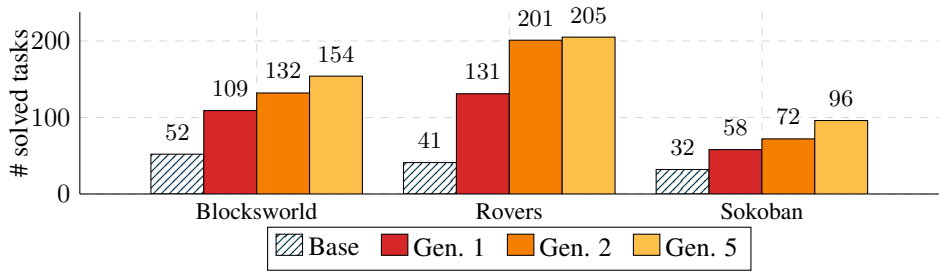

We evaluate this mechanism in the well controlled setting of classical planning.
Iterative deployment captures a pattern common to \emph{planning}: when users use LLMs e.g. to review a product, help solve a reasoning task, or plan a trip, they are more inclined to share LLM results publicly if they are correct. This works as a form of curation, where the users are selecting the correct `solutions' before sharing them. Here, we want to study whether iteratively trained models can improve their planning capabilities having access only to their previously generated curated traces. We simulate the scenario just described in well-controlled environments of classical planning, and focus on the self-improving capabilities of LLMs.
First, we prompt a base
model to solve a diverse set of planning tasks, mixing together both
short-horizon and harder long-horizon planning problems. Then, we discard traces that do not solve the task, mix the remaining into the original data, and fine-tune the next-gen
model. The iteratively trained models essentially
bootstrap each other's planning capabilities: each model attempts to generate solutions to the planning tasks,
without relying on hand-designed prompts or on external planners.  The simple
plans solved in earlier generations are used as part of the training set of
later generations, allowing the model to use these simple ``building blocks'' to
solve more complicated planning problems.

In our experiments using Qwen3 4B \citep{qwen3-tr}, within five (deployment) generations, the
latest model generation achieves more than \emph{double} the base model
performance in all tested domains. Figure~\ref{fig:results-summary} summarizes
the results of our main experiment. In some cases, the performance increases by
a factor of 5. Later generations of the model can also find much longer plans
than the base model, showing that this mechanism allows for out-of-distribution
generalisation. Moreover, there is no significant difference in the average
number of reasoning tokens produced by later generations, contrasting some
results of RL fine-tuning \cite{deepseek-arxiv2025}.

We formally prove that iterative deployment is equivalent to a special case of
REINFORCE \cite{williams-jml1992} when the reward function is binary and traces
are weighted according to importance sampling.
This connection has two important implications: First, it highlights a big safety risk in the deployment of iteratively-trained models, as when the curation is done indirectly through user interactions during post-deployment, the next generation model would effectively be trained with an implicit reward function which can be difficult to control. That could have significant implications on the model's behaviour (e.g.\ the implicit reward could clash with safety training).
Second, the mechanism highlighted here can be viewed as an alternative training regime to explicit RL, preserving generalisation but using a curation mechanism instead of defining an explicit reward
function (which is often challenging for more open-ended, and less
well-specified, tasks). The curation mechanism relies on post-processing validation as a
signal: It works exclusively on the traces generated from prompts provided by
the environment after deployment (e.g., user generated interactions with the
LLM, or tool-usage traces).
Our
observations point the community to study the properties of these implicit reward functions as a
top priority, especially as such iteratively-deployed AI systems are already being used through society.

Throughout the paper we refer to the starting model simply as the base model or as the
\emph{generation 0} of our model. Analogously, we call the method deployed after
$n$ iterations of our process as the \emph{$n$-th generation} of the model.

\section{Iterative Deployment Improves Planning Skills in LLMs}
\label{sec:mechanism}

We study \textbf{Iterative Deployment}, a training mechanism that leads Large Language Models (LLMs) to bootstrap their planning capabilities without requiring external expert demonstrations or additional teacher models. The core intuition is that a model can improve by learning from the simple tasks it has already successfully solved, effectively using its own valid outputs as training data for subsequent generations (provided a reliable curation/verification mechanism is available).

We hypothesize that LLMs improve their planning abilities by simply fine-tuning
over their own curated traces. In other words, they can bootstrap by first
solving easy tasks themselves, and then being fine-tuned using the traces of
these solved tasks. Later generations can then start solving larger tasks, and
these new traces can be used so future generations solve even larger
ones. Repeating this process many times can then gradually improve their planning skills. For
example, if the current generation of a model can solve Sokoban problems with
one single box, then it can learn from its own traces to solve problems with two
boxes, and then with three boxes, etc. So by exploiting its own current
capabilities, the model can improve and solve harder tasks in future
generations.

Formally, let $M_n$ denote the model at generation $n$, parameterized by $\theta_n$. We assume access to a dataset of planning tasks \emph{without} solutions $\mathcal{D}_{tasks}$ and a deterministic external validator $V(x, y)$ which returns true if trace $y$ is a valid solution for task $x$, and false otherwise, with an ability to measure solution efficiency. This validator can be seen as a correction mechanism for reasoning tasks, or even simply as a proxy for user preferences. We evaluate model performance on  $\mathcal{D}_{tasks}$ as a test-set as well. This is because we are interested to find how iterative deployment improves the model's ability to solve longer tasks with access only to its own previously curated solutions. The iterative deployment process then proceeds as follows:

\begin{itemize}
    \item \textbf{Deployment and Trace Collection:} In each iteration $n$, we prompt the current model $M_n$ to solve the tasks in $\mathcal{D}_{tasks}$. For each task input $x$, the model generates a trace $y$ according to its policy $\pi_{\theta_n}(y|x)$. This trace includes the chain-of-thought and the tentative solution generated by the model. This emulates a standard deployment scenario where the model interacts with users or environments.
    \item \textbf{Validation:} The generated traces are passed to the external validator $V$. We filter the outputs to retain only the subset of valid traces, $\mathcal{D}_{valid}^{(n)} = \{(x, y) |V(x, y) = \text{True}\}$. Invalid plans, which certainly constitute the majority of outputs in early generations, are discarded.
    \item \textbf{Curation and Aggregation:} To prevent catastrophic forgetting, reduce model collapse, and further improve generalization, we aggregate the valid traces from the current generation with those from all previous generations. The training dataset for the next step is $\mathcal{T}_{n+1} = \bigcup_{i=0}^{n} \mathcal{D}_{valid}^{(i)}$.
    \item During this aggregation, we apply a second curation step: a \textbf{selection mechanism} to ensure data quality. If multiple valid traces exist for the same task (e.g., from different generations), we retain only the highest-quality solution. In our experiments, quality is defined by plan efficiency -- we select the trace with the shortest plan length, breaking ties by selecting the one with fewer reasoning tokens, but in principle other task-specific metrics can be used.
    \item \textbf{Supervised Fine-Tuning (SFT):} Finally, we produce the next generation $M_{n+1}$ by fine-tuning $M_n$ on the curated dataset $\mathcal{T}_{n+1}$ using the standard supervised learning objective (next token prediction).
\end{itemize}

The curation and aggregation phase is important here, as the likelihood of a
user making an interaction with an LLM publicly available is not uniform, but
depends on the nature of the interaction and the users' intention. For instance,
when a user interacts with an LLM to solve a coding task, they are more likely
to integrate a response that solves their tasks into their codebase. They are
even more likely to use it if the solution is particularly elegant or simple,
akin to our selection mechanism. This creates an effective curation mechanism
controlled by the user's revealed preference (which is not necessarily their
stated preference). This is a key difference to the assumptions used in the study of model collapse \cite{shumailov-et-al-nature2024}.
As LLMs are trained on more and more recursively generated data, their
performance can deteriorate and the models eventually collapse. However, curation might delay or prevent
this.
In this work, we study the effect of this additional assumption compared to the model collapse assumptions, and
argue for the importance of understanding its impact on future model
generations.

Figure~\ref{fig:pipeline} illustrates how one iteration of this process
works. When the $n$-th generation of an LLM is deployed, it is prompted by their users. The produced traces are
filtered by an external \emph{validator}, and those judged \emph{valid traces}
are used to fine-tune the $n+1$-th generation of the model.
Note that we are \emph{not} building a learning curriculum ourselves, but rather the LLM together with the validator are building one: we prompt the model with all tasks from the test set $\mathcal{D}_{tasks}$, discard the invalid traces, add the valid ones to the train set, and then fine-tune the next generation.

\subsection{Formalizing the Connection to Reinforcement Learning}
\label{sec:rl}

Iterative deployment can be interpreted as RL fine-tuning but with the reward
signal left implicit. We prove next that SFT using only valid traces can be seen
as a special case of REINFORCE \cite{williams-jml1992} with an implicit binary reward function.

To show this, we start from the following result:

\begin{proposition}
The update directions of the gradients for SFT using only
valid traces and REINFORCE with binary rewards are identical.
\label{thm:gradient-direction}
\end{proposition}

The proof is included in Appendix~\ref{appendix:proofs}.

In addition to the valid traces generated by the current model, we assume access
to a collection of traces from previous generations of the model. We refer to
the traces produced by current policy $\pi_\theta$ as the on-policy traces, and
those produced by a behavior policy $\pi_\beta$ (earlier generations or any
other external source) as \emph{off-policy} traces.

\begin{proposition}
SFT on a mixture of on- and off-policy valid traces is equivalent to REINFORCE
with binary rewards augmented by importance-weighted contributions from the
behavior policy.
\label{thm:importance-weighted}
\end{proposition}

The proof for Proposition~\ref{thm:importance-weighted} is also included in
Appendix~\ref{appendix:proofs}.

In turn, this proves our original claim that SFT
using only valid traces can be seen as a special case of REINFORCE
\cite{williams-jml1992}.

\begin{proposition}
SFT using only valid traces following the iterative deployment mechanism described in Section~\ref{sec:mechanism} is a special case of REINFORCE with implicitly defined binary rewards.
\label{thm:main-result}
\end{proposition}

\begin{proof}
Follows directly from Proposition~\ref{thm:importance-weighted}.
\end{proof}

\subsection{Implications to AI Safety}

While iterative deployment shares links with RL fine-tuning, it also brings
new concerns about AI safety. Unlike standard RL training, where reward functions can be explicitly
designed to encode human preferences and safety constraints, iterative
deployment of models in the wild relies on implicit signals from post-deployment usage. Curation done indirectly through user interactions with previously deployed models behaves as an opaque reward
function which can be difficult to understand and control.
In RL
for alignment, reward functions are often used to align the model with safety
constraints and goal specifications. In iterative model deployment, the implicit reward
functions could clash with the explicitly specified rewards in
model alignment. This could lead to unexpected problems in future model
generations, as the indirect curation used might lead to large gradient steps in the opposite direction to the gradients induced by the safety training, for example.
Overall, this raises new alignment challenges.

Another concern is bias during validation. If the external validator (e.g., a tool,
or a human curator) has unintended or malicious biases, these biases might
accumulate over the generations. Later models might then optimize for harmful
properties that diverge from the original goals and also from safety
constraints.

A final property to note is model collapse \citep{shumailov-et-al-nature2024}. Data
curation might delay model collapse by filtering for valid traces, it is still
unknown whether this fully prevents collapse. If collapse
occurs, important capabilities can degrade in ways that are difficult to
detect until deployment. Note, however, that it is also unclear whether
longer training periods using RL techniques, such as PPO or GRPO, lead to model
collapse or not.

\section{Empirical Validation}

We evaluate iterative deployment on \emph{classical planning} benchmarks: single-agent problems
with deterministic actions in a fully-observable and discrete environment. We
restrict ourselves to planning domains encoded using PDDL
\citep{mcdermott-et-al-tr1998,haslum-et-al-2019}, which is the description
language used in planning competitions \citep{taitler-et-al-aimag2024}.
Classical planning using LLMs has been widely studied recently in different contexts
\egcite{valmeekam-et-al-neurips2023,stechly-et-al-neurips2024,katz-et-al-neurips2024}. In
\emph{end-to-end planning} with LLMs, the model is prompt with a planning
problem and asked for a solution---called a \emph{plan}---for it. Shorter plans
are usually preferred.

In the context of iterative deployment, we ask the model to solve a set of
planning tasks. Then, we filter the traces that led to valid plans. These valid
traces are then used for the fine-tuning of the next model generation. We include the
valid traces produced by the previous generations as well.

But different model generations might solve the same tasks but with different traces
or different plans. As we reuse valid traces from past generations, this
means that the fine-tuning could contain several traces and plans for the same
input prompt. We decided to have at most one training sample per task during
fine-tuning, and so we use the following rules:
\footnote{
We experimented with alternative variations of these---use several
training samples for the same task, breaking ties randomly---but these
alternatives deteriorated performance. }

\begin{enumerate}
\item if there are several valid traces for the same task, we choose the one that yielded the shortest plan (`simplest', in number of steps)
\item if there is a tie, we select the one with the least amount of reasoning tokens
\end{enumerate}

Additional experimental setup details are given in \S\ref{sec:ExperimentalSetup}.

\paragraph{Benchmark.}

Our benchmark consists of three domains commonly used in classical planning. All
three domains were recently used in the Learning Track of the International
Planning Competition 2023 (IPC 2023) \citep{taitler-et-al-aimag2024}:

\begin{description}[font=\normalfont\itshape]
\item[Blocksworld:]

Several blocks are stacked on a table. The goal is to rearrange them to match a
target configuration. The planning challenge comes from deciding which blocks to
move first and which stacks must be ``destroyed'' in order to reach the target
configuration. Any Blocksworld task can be solved with a simple polynomial
strategy: put all blocks back on the table and then rearrange them according to
the goal. This guarantees that any task with $n$ blocks can be solved within
$2n$ steps.

\item[Rovers:]

This domain is inspired by the 2003 Mars Exploration Rover (MER) missions and
the planned 2009 Mars Science Laboratory (MSL) mission
\cite{long-fox-jair2003}. Rover robots (deployed in Mars) must perform tasks
such as analyzing soil/rock, taking images with cameras (which may need
calibration), and later communicating results-among waypoints and under
visibility constraints. Plan existence for Rovers can be decided in polynomial time \cite{helmert-icaps2006}.

\item[Sokoban:]

This is a PDDL version of the well-known puzzle: an agent in a grid must push
boxes to their goal locations. The grid has several walls and obstacles that may
prevent movement. The agent can only push but never pull boxes. This might lead
to the case where boxes are ``trapped'' in corners which lead to dead-end
states. Sokoban is \pspace-complete \citep{culberson-tr1997}, so plans might be
exponentially long.
\end{description}

\begin{wraptable}{r}{0.55\textwidth}
\vspace{-10pt}
\centering
\begin{tabular}{lrl}
\toprule
\textbf{Domain} & \textbf{Param. Dist.} & \textbf{Explanation} \\ \midrule
Blocksworld     & $n \in [2,10]$        & $n$ blocks              \\
Rovers          & $r \in [1,5]$         & $r$ rovers              \\
Sokoban         & $b \in [1,7]$         & $b$ boxes               \\
\bottomrule                                                       \\
\end{tabular}
\caption{Task distribution for each domain. Showing only the
  main parameters for each domain.}
\label{tab:domain-params}
\vspace{-10pt}
\end{wraptable}

We generated 1000 instances per domain,  varying the difficulty for each instance. This was done by varying specific domain parameters.
Table~\ref{tab:domain-params} explains the
most relevant parameter for each domain and shows their respective
distributions. In this experiment, we train one model per domain; later, we show
results for the case where the generations are fine-tuned over different
domains.

We emphasize that as we are analyzing how the model can self-improve, it is not
relevant whether these domains were already seen during pre-training or not: our
focus is on the relative increase in performance with respect to the base model and not
to external baselines.

\paragraph{Metrics.}

The most relevant metric for our experiments is the number of solved tasks at
each generation. This is our main estimate to know whether new generations are
improving their planning skills or not. We analyze the average number of
solved tasks over 3 complete runs, and also the number of tasks solved
unanimously by all three runs, denoted as unanimous@3. Higher values for
unanimous@3 indicate that the model is more robust, as it is more consistently
able to solve certain tasks despite sampling noise during inference. We are also
interested in the length of the computed plans, which represents the quality of
the solutions, and the length of the reasoning traces, to see if our model's
thinking behavior changes.

\begin{figure}[t]
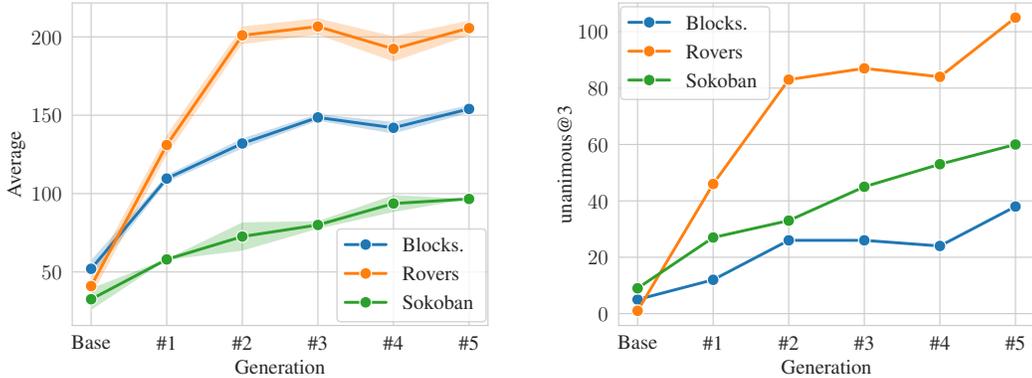

\centering
\begin{subfigure}{0.48\textwidth}
    \centering
    \scalebox{0.55}{\input{coverage-per-gen.pgf}}
    \caption{Average number of solved tasks}
    \label{fig:average}
\end{subfigure}
\hfill
\begin{subfigure}{0.48\textwidth}
    \centering
    \scalebox{0.55}{\input{unanimous-3-per-gen.pgf}}
    \caption{Unanimous@3}
    \label{fig:unanimous-consent}
\end{subfigure}
\caption{Performance of different generations for the different domains tested.}
\label{fig:performance}
\end{figure}

\paragraph{Results.}

Figure~\ref{fig:performance} shows the number solved tasks and unanimous@3 per
domain for each generation. In all three domains, there is a substantial
increase in performance for the first three generations. In later generations,
the speed of improvement decreases. After five generations, the average number of solved tasks in
Blocksworld increased by 196\%, in Rovers by 401\%, and in Sokoban by
196\%. Overall, this indicates that iterative deployment can bring a significant
improvement in planning performance.

In later generations, the number of solved tasks fluctuates slightly between
generations. But this is expected \egcite{deepseek-arxiv2025}, as the
stochasticity in the inference process can cause tasks to be solved (or
unsolved) simply by chance. Overall, still, the trend is clear: the last
generations sustain better performance than earlier ones.

Later generations are also better at the unanimous@3 metric. In particular, the
latest generation achieves the best value for all three domains. This shows that
despite the slow down in performance in the later generations, the models are
still learning to solve some tasks more consistently.

Appendix~\ref{sec:extra-exps} includes more details on these results.

\begin{figure}[t]
\centering
\begin{subfigure}{0.48\textwidth}
    \centering
    \includegraphics[scale=0.3]{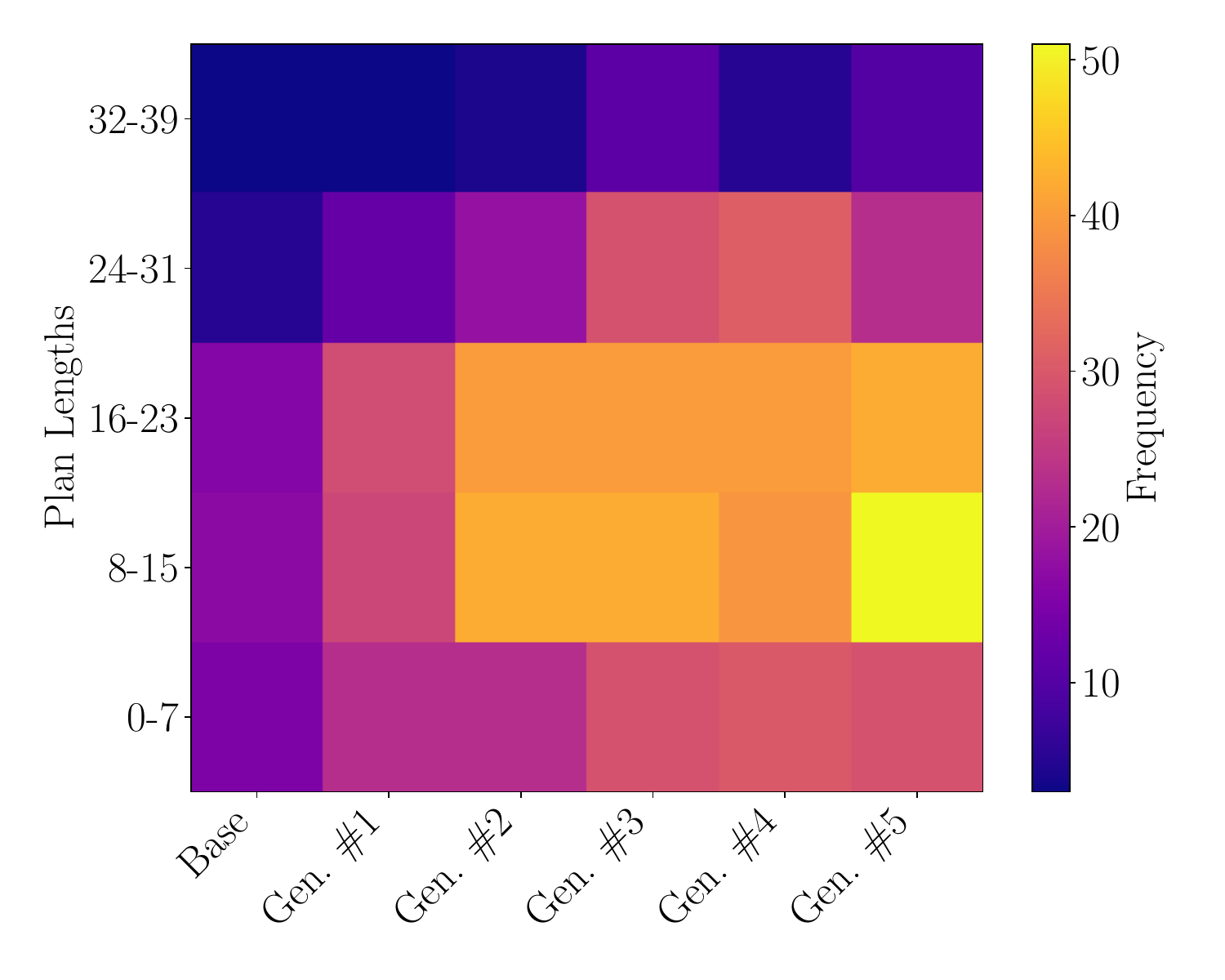}
    \caption{Blocksworld}
    \label{fig:heatmap-bw}
\end{subfigure}
\hfill
\begin{subfigure}{0.48\textwidth}
    \centering
    \includegraphics[scale=0.3]{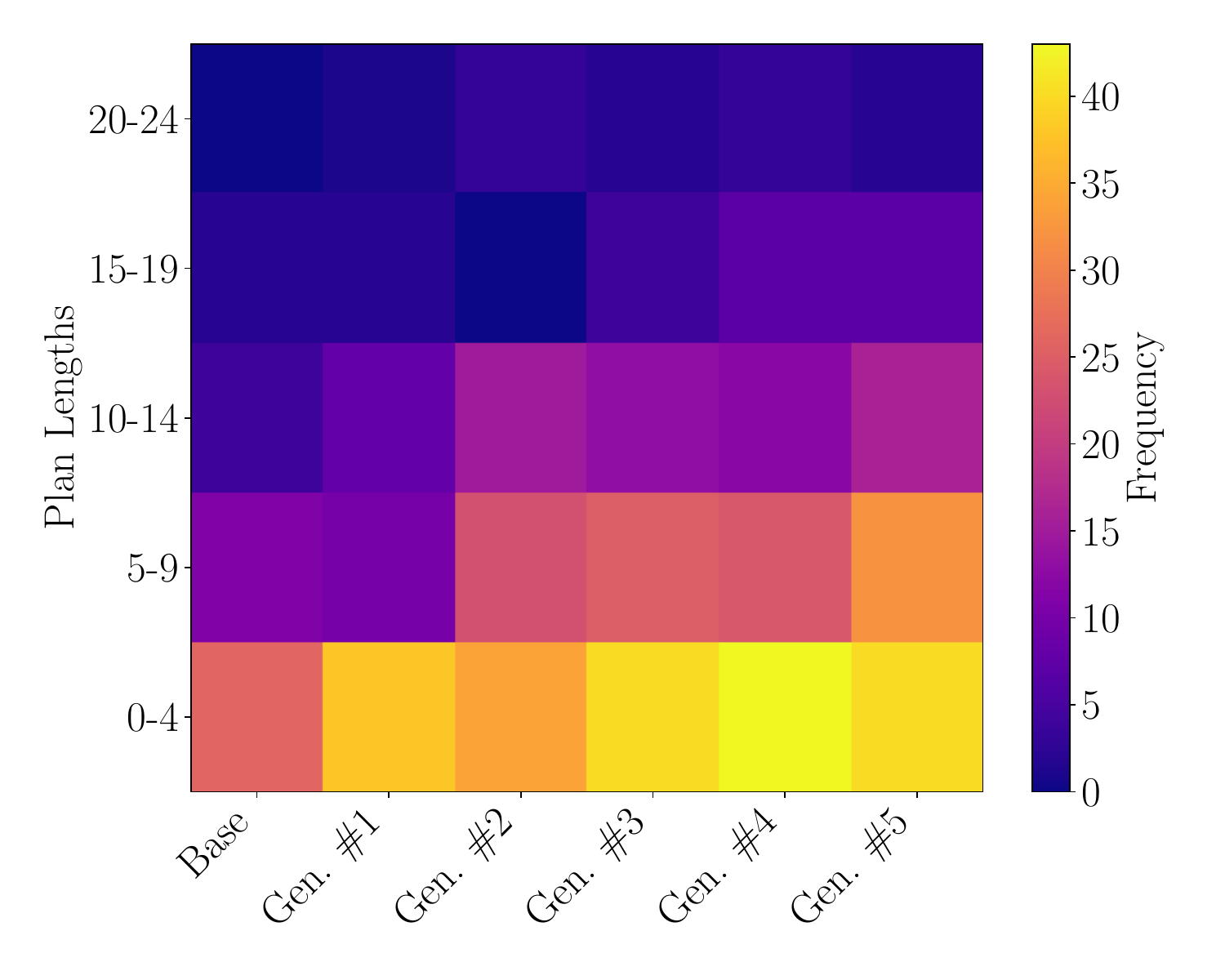}
    \caption{Sokoban}
    \label{fig:heatmap-sokoban}
\end{subfigure}
\caption{Plan length frequency of different generations for the Blocksworld and Sokoban domain.}
\label{fig:selected-domains}
\end{figure}

\paragraph{Is the Model Only Learning Simple Tricks?}
One might wonder whether the models are learning to solve harder tasks or simply learning a few tricks (e.g., format answers correctly, avoid minor mistakes). Our experiments show that later generations are indeed solving longer horizon tasks. Figure~\ref{fig:selected-domains} shows the plan length distribution for different model generations in the Blocksworld and Sokoban domains. It shows the distribution for one single pass over the benchmark. Both figures show a similar trend: as the model evolves, later generations often find longer plans. For example, in Blocksworld, the base model finds mostly plans up to 20 steps, while generation
5 finds many plans up to 35 steps.

Figure~\ref{fig:rovers} shows the cumulative number of solved tasks
per plan length for each generation for the Rovers domain. In this Figure,
outliers are more visible than in Figures~\ref{fig:heatmap-bw} and
\ref{fig:heatmap-sokoban}. We can see that the last three generations find more
plans with 10--30 steps, and generation 5 has significant outliers. However,
when comparing the plans found by two different generations for a same task,
there is no clear trend, and often the plans for a same task have the same
lengths.

\begin{figure}[t]
\centering
\begin{minipage}[t]{0.48\textwidth}
    \centering
    \scalebox{0.75}{\input{rovers-plot.pgf}}
    \caption{Performance across generations for Rovers. The $y$-axis represents
      the plan length of the plans found by each generation, while the $x$-axis
      shows the cumulative number of solved instances sorted by plan length.}
    \label{fig:rovers}
\end{minipage}
\hfill
\begin{minipage}[t]{0.48\textwidth}
    \centering
    \scalebox{0.575}{\input{token-curves.pgf}}
    \caption{Average number of reasoning tokens per generation for each
      domain. Shaded area corresponds to one standard deviation of the
      mean. Considering both valid and invalid traces.}
    \label{fig:tokens}
\end{minipage}
\end{figure}

The last metric studied for this experiment is the amount of reasoning tokens
produces by each generation. In RL fine-tuning, the model often increases the
number of reasoning tokens produced as training progresses
\citep{deepseek-arxiv2025}. Figure~\ref{fig:tokens} shows that this does not happen
in our case. We average over both valid and invalid traces in this analysis. For
Blocksworld and Sokoban, later generations have slightly fewer reasoning tokens
in average, while for Rovers the result is the opposite. However, the difference
between the base model and the last tested generation is not so significant:
around 2000 tokens difference per domain. If we restrict the data to the valid
traces, we still observe the same behaviour.

\paragraph{Does Curation Really Matter?}

\begin{wraptable}{r}{0.55\textwidth}
\vspace{-10pt}
\centering
\begin{tabular}{lrl}
\toprule
\textbf{Gen.} & \textbf{w/ Curation}     & \textbf{w/o Curation}   \\ \midrule
Base          & \textbf{52.0 $\pm$ 6.3}  & \textbf{52.0 $\pm$ 6.3} \\
Gen. 1        & \textbf{109.6 $\pm$ 2.8} & 71.0 $\pm$ 3.2          \\
Gen. 2        & \textbf{132.0 $\pm$ 3.7} & 82.0 $\pm$ 4.1          \\
Gen. 3        & \textbf{148.6 $\pm$ 2.6} & 84.3 $\pm$ 4.9          \\
Gen. 4        & \textbf{142.0 $\pm$ 4.2} & 84.0 $\pm$ 2.1          \\
Gen. 5        & \textbf{154.0 $\pm$ 2.9} & 79.3 $\pm$ 5.2          \\
\bottomrule                                                        \\
\end{tabular}
\caption{Average number of solved tasks in Blocksworld for iterative deployment
  with and without curation.
}
\label{tab:no-curation}
\vspace{-1pt}
\end{wraptable}

We also studied how curation impacts our results. To analyze that, we repeated
the experiment for the Blocksworld domain but \emph{without} curation. In other words,
we used all traces from previous generations to fine-tune the next one. (We do
not include traces that reached the maximum token generation limit though.)
Table~\ref{tab:no-curation} compares the average number of solved tasks for the
case with and without curation. The version without curation also improves on
the base model. While this might seem surprising at first, this is still
expected: with more data, the version without curation is still able to acquire
minor improvements, such as a better understanding of the PDDL syntax and more
reliability at following instructions.
The version with curation has a clear edge on the version without curation. This
shows that, as we expected, the curation mechanism is crucial to achieve a
significant improvement using iterative deployment.
Furthermore, the version with curation requires less data. For example, to train
the 5th generation of each model, the version with curation only used 356
traces, while the version with curation used 4017. So curation is capable of
improving performance by 94\% while using a fraction of the training data.

\section{Related Work}

\paragraph{Reasoning with LLMs.}

Chain-of-Thought (CoT) \citep{wei-et-al-neurips2022} uses ``reasoning tokens''
to improve the performance of LLMs in many problems.  Many approaches build on
this idea \egcite{yao-et-al-neurips2023,sel2024algorithm}s. For example,
different methods explore multiple reasoning paths to find a valid
solution~\citep{yao-et-al-neurips2023,snell-et-al-icrl2025}. Others show that
using more informative reasoning traces, such as those based on action sequence
similarity, can also enhance planning performance \citep{zhao-et-al-icrl2025}.
However, these approaches often have limited success in generalizing to
out-of-distribution reasoning tasks and longer-horizon planning problems. For
instance, \citet{stechly-et-al-neurips2024} show that while CoT can help with
planning problems, it does not consistently lead to generalization, particularly
for out-of-distribution tasks.

\paragraph{Improving Reasoning Capabilities of LLMs.}

There are several methods in the literature that try to improve the reasoning
capacities of LLMs
\egcite{zhang-et-al-neurips2024,muennighoff-et-al-arxiv2025,wen-et-al-arxiv2025}. The
iterative deployment mechanism we studied here shares conceptual similarities
with the Self-Taught Reasoner (STaR) framework
\cite{zelikman-et-al-neurips2022}, which iteratively improves a model's
reasoning ability by fine-tuning on traces that lead to valid answers.
However, while STaR tries to improve model performance, we focus on a very
different question: we want to understand whether performance improvement could
emerge \emph{unintentionally} from repeated model deployments. We use planning
problems to ground this question in a controlled setting which allows us to
study this phenomenon, and prove that repeated deployment can be seen as a form
of RL. On the technical level, there are key differences between the two as
well.  First, STaR has an extra-step for producing \emph{rationales} for
wrong answers, which are often not available from scraped user-curated
traces. Second, we use valid traces from older generations as well as the
current one emulating the property of web-scraped data containing traces from
multiple former model generations, while STaR focuses on traces from the current
model generation.

Reinforcement learning has been widely used to improve the reasoning
capabilities of large language models
\egcite{shao-et-al-arxiv2024,deepseek-arxiv2025}. Techniques such as group
relative optimization (GRPO) allow us to fine-tune models without supervised
data, using only an internal reward function \cite{shao-et-al-arxiv2024}. By
correctly modeling the reward functions, we can also influence the RL training
so that the models better align with safety conditions or goal specifications.

Iterative deployment is also related to test-time scaling methods. For example,
\citet{muennighoff-et-al-arxiv2025} propose a simple test-time scaling technique
where they fine-tune a small model on traces generated by a much larger, more
capable teacher model---in their paper, the small model is a 32B model, while
the teacher model is a model from the Gemini family. In contrast, iterative deployment
does not rely on a separate, more powerful model. Instead, the LLM generates its
own training data, bootstrapping its performance by solving progressively harder
tasks. The model effectively becomes its own teacher, curating its own
experience to improve in the next generations.

\paragraph{Model Collapse.}
One important connection to our analysis is `model collapse'. \citet{shumailov-et-al-nature2024} show that iteratively training
models with their own synthetic output eventually collapses (i.e., the model's
distribution shrinks so its tails disappear). The key difference in our analysis
is the explicit curation step: only valid traces, as determined by an external
validator, are used for fine-tuning.
\citet{shumailov-et-al-nature2024} lacks a curation step, and assumes all data is kept between generations.
We show that curation can improve reasoning, but it is unknown whether the curation step completely avoids, or simply delays, model collapse. In our context, we have
tested running our experiments above up to 10 generations and (despite the
smaller improvement past the fifth generation) but we did not observe imminent
hints of model collapse within planning domain.

\paragraph{Planning using LLMs.}

\citet{valmeekam-et-al-neurips2023} show that early LLMs do not reliably solve
simple planning tasks. \citet{kokel-et-al-aaai2025} show that problems simpler
than computing a plan, such as verifying if a given plan is valid, are already
challenging for LLMs.

In the context of classical planning, several fine-tuning strategies have been
explored. \citet{rossetti-et-al-icaps2024} train domain-specific GPT models for
each planning domain, achieving strong performance but only for in-distribution
problems. \citet{bohnet-et-al-arxiv2024} report that supervised fine-tuning with
optimal plans produced from an off-the-shelf planner \citep{helmert-jair2006}
did not lead to out-of-distribution
generalization. \citet{huang-et-al-icaps2025} apply RL to end-to-end plan
generation, but observed only a small performance improvement. Our work differs
from these, by showing that generalization can be achieved without access to an
existing planner or to an explicit reward modeling, relying instead on an
iterative process with a simple validation filter.

\citet{verma-et-al-arxiv2025} propose an instruction-tuning framework that
improves symbolic planning by teaching LLMs explicit logical CoT reasoning with
feedback from VAL. While conceptually related to the iterative deployment
mechanism, their method relies on carefully handcrafted training data and a
multi-phase fine-tuning using structured reasoning feedback from a validator,
whereas the mechanism studied here achieves self-improvement through repeated
deployment and fine-tuning solely on valid traces, requiring just a binary
signal.

\section{Conclusion}

In this work, we showed that iterative deployment of LLMs, where each generation
is fine-tuned on curated traces from previous deployments, improves planning
capabilities. Our experimental results on classical planning domains show that
models trained under this paradigm more than doubled their performance within
five generations, with evidence of out-of-distribution generalization to
longer-horizon plans.

Iterative deployment can be an effective alternative to RL to improve reasoning
capabilities of LLMs. We proved theoretically that our method is equivalent to a
special case of REINFORCE with binary rewards. Unlike RL fine-tuning, our
approach does not rely on carefully designed reward functions. Instead, it uses
external postprocessing tools to validate/curate LLM answers as an implicit
signal.  However, the absence of an internal reward function brings concern
about potential biases and unintended effects caused during training. The use of
external validators, that might be outside our control, can be damaging to AI
safety.

Overall, our results suggest that iterative deployment is a viable method for
self-improving LLMs. In future work, we plan to study how model collapsing
\citep{shumailov-et-al-nature2024} can affect iterative deployment, and also to
develop theoretical results to connect our method with RL fine-tuning more
explicitly.

\bibliographystyle{plainnat}
\bibliography{abbrv,literatur,additional,crossref}

\newpage
\appendix
\section{Appendix}

\subsection{Proofs}
\label{appendix:proofs}

In this section we present the proofs for our the theoretical results of
Section~\ref{sec:rl}. We use the same notation and terminology from the paper.

\setcounter{proposition}{0}

\begin{proposition}
The update directions of the gradients for SFT using only
valid traces and REINFORCE with binary rewards are identical.
\end{proposition}

\begin{proof}
Let $x$ denote an input (e.g., a prompt), and let $y = (y_1,\dots,y_T)$ denote a
complete trace produced by a policy $\pi_\theta(y \mid x)$ with parameters
$\theta$ (e.g.\ the LLM above).

We assume a binary reward function $R(x,y) \in \{0,1\}$, where $R(x,y) = 1$ if
$y$ is a valid trace for $x$, and 0 otherwise.

The learning objective of REINFORCE is the expected reward:
\begin{align*}
J(\theta) = \mathbb{E}_{x \sim \mu} \, \mathbb{E}_{y \sim \pi_\theta(\cdot\mid x)}[R(x,y)].
\end{align*}
The gradient $\nabla_\theta J(\theta)$ is
\begin{align*}
\nabla_\theta J(\theta)
= \mathbb{E}_{x,y}
\big[ R(x,y)\,\nabla_\theta \log \pi_\theta(y\mid x) \big].
\end{align*}

Using the factorization
\[
\log \pi_\theta(y\mid x)
= \sum_{t=1}^T \log \pi_\theta(y_t \mid y_{<t}, x),
\]
we obtain
\[
\nabla_\theta \log \pi_\theta(y\mid x)
= \sum_{t=1}^T
\nabla_\theta \log \pi_\theta(y_t \mid y_{<t}, x).
\]

Thus, we can rewrite $\nabla_\theta J(\theta)$ as
\[
\nabla_\theta J(\theta)
=
\mathbb{E}_{x,y}
\left[
R(x,y) \sum_{t=1}^T
\nabla_\theta
\log \pi_\theta(y_t \mid y_{<t}, x)
\right].
\]

Suppose we sample a dataset $\mathcal{D} = \{(x_i, y_i, R_i)\}_{i=1}^N$ for $N \in \mathbb{N}$ with $R_i \in \{0,1\}$.
A Monte Carlo estimate of the REINFORCE gradient is
\begin{align*}
\widehat{\nabla_\theta J} = \frac{1}{N} \sum_{i=1}^N R_i \sum_{t} \nabla_\theta \log \pi_\theta(y_{i,t} \mid y_{i,<t}, x_i).
\end{align*}

Let $\mathcal{D}_+ = \{(x_i,y_i) : R_i = 1\}$ be the set of valid traces, and $\mathcal{D}_- = \{(x_i,y_i) : R_i = 0\}$ be the set of invalid traces.
As our reward function $R(x,y)$ is binary, invalid traces do not affect the gradient (i.e., they contribute zero). Therefore, we can simplify the gradient:
\begin{align*}
\widehat{\nabla_\theta J} =
\frac{1}{N} \sum_{i \in \mathcal{D}_+}
\sum_t \nabla_\theta
\log \pi_\theta(y_{i,t} \mid y_{i,<t}, x_i).
\end{align*}

Let $N_+ = |\mathcal{D}_+|$, then
\begin{align}
\widehat{\nabla_\theta J}
=
\frac{N_+}{N}
\left[
\frac{1}{N_+}
\sum_{i \in \mathcal{D}_+}
\sum_t \nabla_\theta \log \pi_\theta(y_{i,t}\mid y_{i,<t},x_i)
\right].
\label{eq:gradient-reinforce}
\end{align}

On the flip side, SFT the valid traces uses the loss
\[
\mathcal{L}_{\text{SFT}}(\theta)
=
- \frac{1}{N_+}
\sum_{i\in \mathcal{D}_+}
\sum_{t=1}^{T_i}
\log \pi_\theta(y_{i,t} \mid y_{i,<t}, x_i).
\]

Its gradient is
\[
\nabla_\theta \mathcal{L}_{\text{SFT}}(\theta)
=
-
\frac{1}{N_+}
\sum_{i\in \mathcal{D}_+}
\sum_t
\nabla_\theta \log \pi_\theta(y_{i,t} \mid y_{i,<t}, x_i).
\]

We can then rewrite \eqref{eq:gradient-reinforce} in terms of $\nabla_\theta
\mathcal{L}_{\text{SFT}}(\theta)$:
\[
\widehat{\nabla_\theta J}
=
- \frac{N_+}{N} \,
\nabla_\theta \mathcal{L}_{\text{SFT}}(\theta).
\]

Thus, a gradient ascent step on $J(\theta)$,
\[
\theta \leftarrow
\theta + \eta \widehat{\nabla_\theta J},
\]
is equivalent (up to the positive scaling factor $N_+/N$) to a gradient
descent step on $\mathcal{L}_{\text{SFT}}$:
\[
\theta \leftarrow
\theta - \bigl(\eta\,N_+/N \bigr)
\nabla_\theta \mathcal{L}_{\text{SFT}}(\theta).
\]

Hence, the update directions are identical.
\end{proof}

\begin{proposition}
SFT on a mixture of on- and off-policy valid traces is equivalent to REINFORCE
with binary rewards augmented by importance-weighted contributions from the
behavior policy.
\end{proposition}
\begin{proof}
Let $p_\pi^+(x,y)$ denote the distribution over valid traces induced by
sampling $x \sim \mu(x)$, then sampling $y \sim \pi_\theta(\cdot \mid x)$, and
retaining the pair only if $R(x,y)=1$:
\[
p_\pi^+(x,y)
\;\propto\;
\mu(x)\,\pi_\theta(y\mid x)\,\mathbf{1}\{R(x,y)=1\}.
\]
Similarly, let
\[
p_\beta^+(x,y)
\;\propto\;
\mu(x)\,\pi_\beta(y\mid x)\,\mathbf{1}\{R(x,y)=1\}
\]
denote the distribution over valid off-policy traces.

The dataset of valid traces can be modeled as a mixture
\[
p_{\mathrm{data}}(x,y)
=
(1-\lambda)\,p_\pi^+(x,y)
+ \lambda\,p_\beta^+(x,y),
\qquad \lambda \in [0,1],
\]
where $\lambda$ reflects the proportion of off-policy data.

We can then show that SFT on a mixture of on- and off-policy valid traces
is equivalent to REINFORCE with binary rewards augmented by importance-weighted
contributions from the behavior policy.

The SFT objective over all valid trace is
\[
\mathcal{L}_{\mathrm{SFT}}(\theta)
=
-\mathbb{E}_{(x,y)\sim p_{\mathrm{data}}}
\big[\log \pi_\theta(y\mid x)\big].
\]

Using the mixture form of $p_{\mathrm{data}}$, this decomposes as
\[
\mathcal{L}_{\mathrm{SFT}}(\theta)
=
-(1-\lambda)\,\mathbb{E}_{p_\pi^+}
\big[\log \pi_\theta(y\mid x)\big]
-
\lambda\,\mathbb{E}_{p_\beta^+}
\big[\log \pi_\theta(y\mid x)\big].
\]

Differentiating $\mathcal{L}_{\mathrm{SFT}}(\theta)$ we obtain the following decomposition of the gradient:
\[
\nabla_\theta \mathcal{L}_{\mathrm{SFT}}(\theta)
=
-(1-\lambda)\,\mathbb{E}_{p_\pi^+}
\big[\nabla_\theta \log \pi_\theta(y\mid x)\big]
-
\lambda\,\mathbb{E}_{p_\beta^+}
\big[\nabla_\theta \log \pi_\theta(y\mid x)\big].
\]

The first term, corresponding to the on-policy traces, is identical (up to a
scalar factor) to REINFORCE with binary reward $R(x,y)\in\{0,1\}$ under the
current policy:
\[
\mathbb{E}_{p_\pi^+}
\big[\nabla_\theta \log \pi_\theta(y\mid x)\big]
\;\propto\;
\mathbb{E}_{x\sim \mu,\, y\sim \pi_\theta}
\big[R(x,y)\,\nabla_\theta \log \pi_\theta(y\mid x)\big].
\]

To express the second term, corresponding to the off-policy traces, as an
expectation under the current policy, we apply importance sampling
\cite{precup-et-al-icml2000}. For any valid trace $(x,y)$,
\[
\frac{p_\beta^+(x,y)}{p_\pi^+(x,y)}
=
\frac{\pi_\beta(y\mid x)}{\pi_\theta(y\mid x)},
\]
so
\[
\mathbb{E}_{p_\beta^+}
\big[\nabla_\theta \log \pi_\theta(y\mid x)\big]
=
\mathbb{E}_{p_\pi^+}
\left[
\frac{\pi_\beta(y\mid x)}{\pi_\theta(y\mid x)}\,
\nabla_\theta \log \pi_\theta(y\mid x)
\right].
\]

Substituting this expression produces the fully on-policy-looking form
\[
\nabla_\theta \mathcal{L}_{\mathrm{SFT}}(\theta)
=
-\mathbb{E}_{p_\pi^+}
\Big[
\big((1-\lambda)
+
\lambda\,\tfrac{\pi_\beta(y\mid x)}{\pi_\theta(y\mid x)}\big)
\,\nabla_\theta \log \pi_\theta(y\mid x)
\Big].
\]

Thus the SFT update may be interpreted as a REINFORCE-like gradient step under
the current policy, restricted to valid trace, with an \emph{effective
reward}
\[
R_{\mathrm{eff}}(x,y)
\;\propto\;
(1-\lambda)
+
\lambda\,\frac{\pi_\beta(y\mid x)}{\pi_\theta(y\mid x)}.
\]

This shows that SFT on a mixture of on- and off-policy valid traces corresponds
to REINFORCE with binary rewards augmented by importance-weighted contributions
from the behavior policy. In turn, this proves our original claim that SFT
using only valid traces can be seen as a special case of REINFORCE
\cite{williams-jml1992}.
\end{proof}

\subsection{Experimental Setup}
\label{sec:ExperimentalSetup}
All our experiments were run on Intel Xeon Gold 6248R processors running at
3.00 GHz with an NVIDIA A100 80 GiB GPU.
We use the Qwen3 4B as the base model \citep{qwen3-tr}. Specifically, we use the
Qwen3 4B Thinking 2507 version, which was better at following formatting
instructions during our early experiments. We use temperature $0.6$ and a
maximum context length of 32\,768 tokens during inference.
To fine-tune our models, we use low-rank adaptation (LoRA)
\citep{hu-et-al-iclr2022}. The hyperparameters used for fine-tuning can be found
in Appendix~\ref{sec:hyperparameters}. To avoid stacking up or retraining existing adapters,
when we train generation $n$, we fine-tune the base model Qwen3 4B with the
traces from generations $0, 1, 2, \dots, n-1$. To validate the computed plans, we use
VAL \citep{howey-long-icaps2003wscompetition}, which is commonly used in
planning competitions.

\paragraph{Prompt.}

Our prompt format is based on the prompt used by \citet{correa-et-al-neurips2025}
for end-to-end PDDL planning. To solve a task $T$ from a domain $D$, the
prompt contains the following components:

\begin{enumerate}
\item The following instruction: ``\textit{You are generating plans for PDDL tasks. You
  will be given the PDDL domain and the PDDL instance, and you need to return
  the plan.}''
\item Two examples of PDDL domains, tasks, and their respective plans.
\item The PDDL domain $D$ and the PDDL task $T$.
\end{enumerate}
We use tasks from the traditional Gripper and Logistics domains as
examples. These domains are not included in our experiments. The plans used
to illustrate the examples are not optimal plans.

\subsection{Fine-Tuning Hyperparameters}
\label{sec:hyperparameters}

Below we include the hyperparameters used during our supervised fine-tuning with
LoRA. At each generation, we randomly select 10$\%$ of the traces to be used as
validation set. The validation loss is evaluation every 10 steps.

\begin{center}
\begin{tabular}{lr}
\toprule
\textbf{Hyperparameter}      & \textbf{Value}     \\ \midrule
 Maximum Context Length      & 32\,768            \\
 Batch Size                  & 1                  \\
 Gradient Accumulation Steps & 1                  \\
 Epochs                      & 2                  \\
 Optimizer                   & AdamW (fused)      \\
 Learning Rate               & $1 \times 10^{-5}$ \\
 Learning Rate Scheduler     & Cosine             \\
 Warm-up Ratio                & 0.02               \\
 Weight Decay                & 0.01               \\
 Max. Gradient Norm          & 1.0                \\
 LoRA Rank                   & 16                 \\
 LoRA Alpha                  & 32                 \\
 LoRA Dropout                & 0.05               \\
 LoRA Bias                    & None               \\
\bottomrule
\end{tabular}
\end{center}

\subsection{Additional Results}
\label{sec:extra-exps}

 \begin{table}[t!]
\begin{center}
\resizebox{\textwidth}{!}{
\begin{tabular}{lrrrrrrrrrrcr}
\toprule
Domain      & Base           & \#1             & \#2             & \#3                      & \#4               & \#5                      \\
\midrule
Blocksworld & 52.0 $\pm$ 6.3 & 109.6 $\pm$ 2.8 & 132.0 $\pm$ 3.7 & 148.6 $\pm$ 2.6          & 142.0 $\pm$ 4.2   & \textbf{154.0 $\pm$ 2.9} \\
Rovers      & 41.0 $\pm$ 7.7 & 131.0 $\pm$ 7.2 & 201.0 $\pm$ 6.1 & \textbf{206.6 $\pm$ 5.7} & 192.33 $\pm$  8.5 & 205.6  $\pm$ 5.2         \\
Sokoban     & 32.6 $\pm$ 7.4 & 58.0 $\pm$ 0.8  & 72.6  $\pm$ 9.5 & 80.0 $\pm$ 2.9           & 93.67 $\pm$  5.9  & \textbf{96.6 $\pm$ 0.4}  \\
\bottomrule                                                                                                                                \\
\end{tabular}
}
\caption{Average number of solved tasks per generation of the model. Average over three runs. Total of 1000 tasks per domain. Best value for each domain is highlighted.}
\label{tab:solved}
\end{center}
\end{table}

\begin{table}[t!]
\begin{center}
\begin{tabular}{lrrrrrrrrrrcr}
\toprule
Domain             & Base & \#1 & \#2 & \#3 & \#4 & \#5          \\
\midrule
Blocksworld (1000) & 5    & 12  & 26  & 26  & 24  & \textbf{38}  \\
Rovers (1000)      & 1    & 46  & 83  & 87  & 84  & \textbf{105} \\
Sokoban (1000)     & 9    & 27  & 33  & 45  & 53  & \textbf{60}  \\
\bottomrule                                                      \\
\end{tabular}
\caption{Number of solved tasks in unanimous@3 per generation of the model. Best value for each
  domain is highlighted.}
\label{tab:unanimousn}
\end{center}
\end{table}

Table~\ref{tab:solved} shows the number solved tasks per domain for each
generation of our process. Table~\ref{tab:unanimousn} shows the unanimous@3 for
each generation.

\subsection{Other IPC 2023 Domains}
\label{sec:domains}

We briefly explain the seven domains used in our second experiment with 10
different domains. The other three domains --- Blocksworld, Rovers, and Sokoban
--- where explained in the text. Table~\ref{tab:domain-params-small-exp} shows
the distribution of the main parameters for each domain used in this experiment.

\paragraph{Childsnack}

This domain involves making a number of sandwiches, some with gluten‐free
ingredients for children with allergies, placing them on trays, and delivering
them to tables. Key constraints include production (gluten and non‐gluten
items), packaging (quantity of trays), and serving order.

\paragraph{Ferry}

Cars are distributed across locations, and a ferry (which can fit one car) must
transport them to their target destinations. The domain requires deciding the
ferry’s schedule, routing, and when to move each car, all under capacity
constraints.

\paragraph{Floortile}

A grid of tiles must be painted by robots that can move in four directions via
unpainted tiles. Robots can only move over tiles that are not yet painted, so
they must paint tiles in a certain order.

\paragraph{Miconic}

Elevator domain: there are multiple floors, passengers on origin floors, and
destinations; an elevator must move and pick up / drop off passengers,
respecting boarding and alighting constraints. Though conceptually simple, it
tests planner efficiency regarding sequencing and movement decisions, especially
as passenger numbers and floor counts increase.

\paragraph{Satellite}

Satellites have instruments (which may need calibration) and different modes of
observation; they must orient, turn, switch instruments, and gather images to
satisfy a given goal. Only one instrument can be active at a time, and switching
or calibrating requires actions.

\paragraph{Spanner}

An agent traverses a path where there are $k$ spanners place along this path. At
the end of the path there exists a gate with $m \leq k$ loose nuts. The agent
must collect enough spanners to tighten all nuts and cross corridors to be able
to cross the gate. Spanners break when used, thus the agent must collect enough
spanners to open the gate.

\paragraph{Transport}

This is a traditional logistics problem with capacity constraints. Vehicles must
deliver packages to certain locations on a graph, subject to vehicle capacities
(i.e., number of packages that can be transported at once).

\begin{table}[t!]
\centering
\begin{tabular}{lrl}
\toprule
\textbf{Domain} & \textbf{Param. Dist.} & \textbf{Param. Meaning}               \\ \midrule
Blocksworld     & $n \in [2,10]$        & $n$ blocks to be rearranged           \\
Childsnack      & $c \in [4,12]$        & $c$ children to be served             \\
Ferry           & $a \in [3,30]$        & $a$ cars to be transported            \\
Floortile       & $t \in [2,25]$        & $t$ tiles to paint                    \\
Miconic         & $p \in [1, 10]$       & $p$ passengers to deliver             \\
Rovers          & $r \in [1,4]$         & $r$ rovers to drive and collect items \\
Satellite       & $s \in [1,8]$         & $s$ satellites to synchronize         \\
Sokoban         & $b \in [1,4]$         & $b$ boxes to push to their goals      \\
Spanner         & $k \in [1,20]$        & $k$ spanners to collect               \\
Transport       & $v \in [2,10]$        & $v$ vehicles to drive                 \\
\bottomrule \\
\end{tabular}
\caption{Task distribution and plan lengths for each domain. Showing only the
  main parameter for each domain.}
\label{tab:domain-params-small-exp}
\end{table}

\end{document}